%% file: main.tex
\documentclass[10pt,onecolumn,letterpaper]{article}

\usepackage[pagenumbers]{cvpr} 

\input{math_commands.tex}

\input{preamble}

\definecolor{cvprblue}{rgb}{0.21,0.49,0.74}
\usepackage[pagebackref,breaklinks,colorlinks,citecolor=cvprblue]{hyperref}

\def\paperID{*****} 
\def\confName{CVPR}
\def\confYear{2024}

\title{CCM: Adding Conditional Controls to Text-to-Image Consistency Models}

\author{
Jie Xiao$^{1}$,
Kai Zhu$^2$,
Han Zhang$^3$,
Zhiheng Liu$^1$,
Yujun Shen$^4$,
Yu Liu$^2$,
Xueyang Fu$^1$,
Zheng-Jun Zha$^1$ \\[5pt]
$^1$USTC \qquad
$^2$Alibaba Group \qquad
$^3$SJTU \qquad
$^4$Ant Group
}

\begin{document}
\maketitle
\input{sec/0_abstract}    
\input{sec/1_intro}
\input{sec/2_method}
\input{sec/3_exp}
\input{sec/4_relate_work}
\input{sec/5_conclus}
{
    \small
    \bibliographystyle{ieeenat_fullname}
    \bibliography{main}
}


\end{document}

%% file: math_commands.tex
\usepackage{amsmath,amsfonts,bm}

\def\eqref#1{equation~\ref{#1}}

\def\1{\bm{1}}

\def\vc{{\bm{c}}}

\def\vf{{\bm{f}}}

\def\vx{{\bm{x}}}
\def\vy{{\bm{y}}}

\DeclareMathAlphabet{\mathsfit}{\encodingdefault}{\sfdefault}{m}{sl}
\SetMathAlphabet{\mathsfit}{bold}{\encodingdefault}{\sfdefault}{bx}{n}

\def\gL{{\mathcal{L}}}

\def\gN{{\mathcal{N}}}

\def\gU{{\mathcal{U}}}

\newcommand{\pdata}{p_{\rm{data}}}

\newcommand{\E}{\mathbb{E}}

\DeclareMathOperator*{\argmin}{arg\,min}

%% file: preamble.tex
\usepackage[dvipsnames]{xcolor}


%% file: sec/0_abstract.tex
\begin{abstract}
Consistency Models (CMs) have showed a promise in creating visual content efficiently and with high quality. However, the way to add new conditional controls to the pretrained CMs has not been explored. In this technical report, we consider alternative strategies for adding ControlNet-like conditional control to CMs and present three significant findings. 1) ControlNet trained for diffusion models (DMs) can be directly applied to CMs for high-level semantic controls but struggles with low-level detail and realism control. 2) CMs serve as an independent class of generative models, based on which ControlNet can be trained from scratch using Consistency Training proposed by Song et al~\cite{song2023consistency,song2023improved}. 3) A lightweight adapter can be jointly optimized under multiple conditions through Consistency Training, allowing for the swift transfer of DMs-based ControlNet to CMs.
We study these three solutions across various conditional controls, including edge, depth, human pose, low-resolution image and masked image with text-to-image latent consistency models. Project page: \url{https://swiftforce.github.io/CCM}. 
\end{abstract}

%% file: sec/1_intro.tex
\begin{figure}[!htb]
    \centering
    \includegraphics[width=\linewidth]{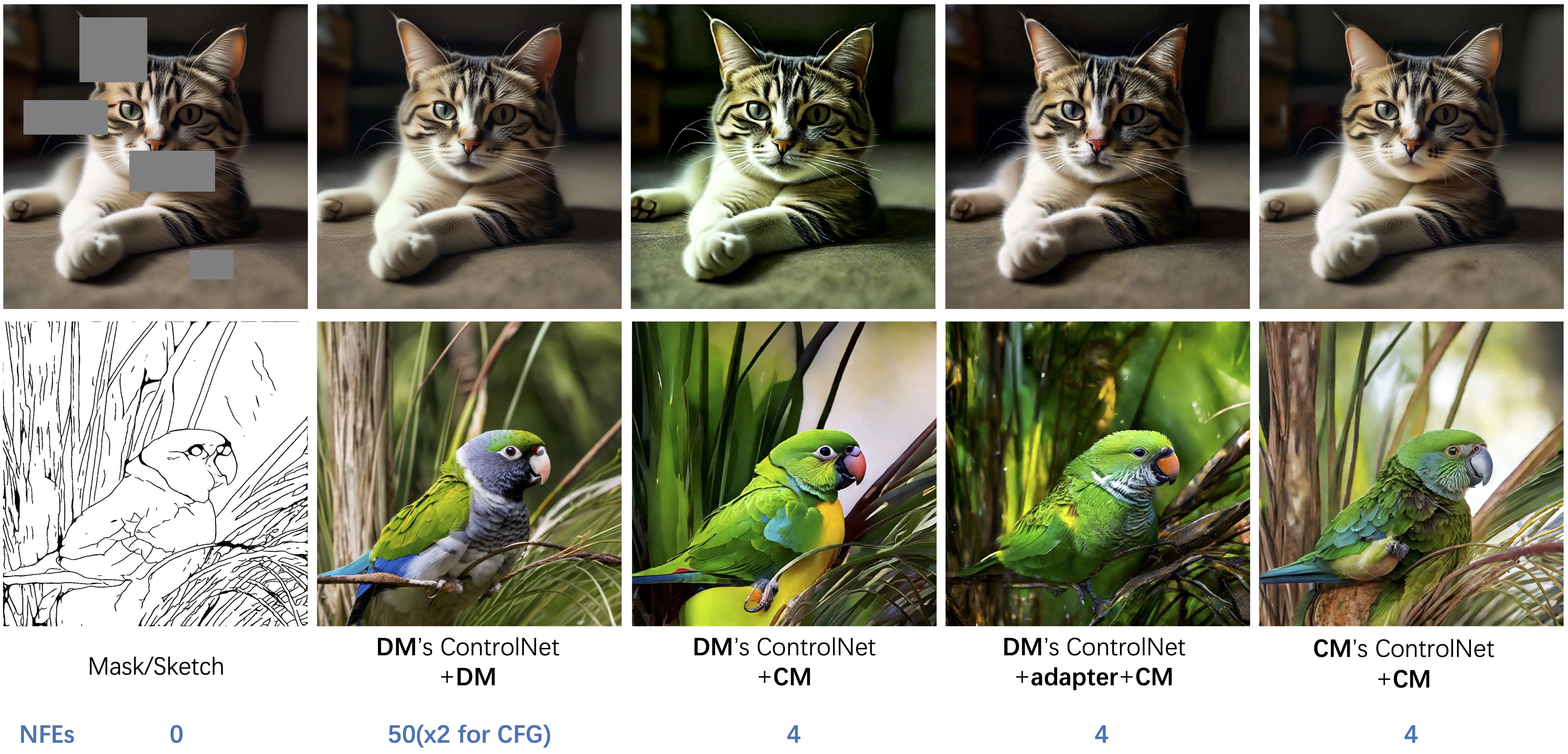}
    \caption{Visual comparison of different strategies of adding controls at $1024$x$1024$ resolution. NFEs: the number of function evaluations; CFG: classifier free guidance.}
    \label{fig:ctrl_maincompare}
\end{figure}

\begin{table}[!htbp]
\centering
\caption{Summary of symbols.}
\label{tab:symbol}
\begin{center}
\begin{tabular}{lc}
\hline
    $\bm{\phi}$ & trainable paramaters of diffusion model\\
    $\bm{\theta}$ & trainable paramaters of consistency model \\
    $\bm{\psi}$ & trainable paramaters of ControlNet \\
    $\bm{\Delta\psi}$ & trainable paramaters of adapter \\
    $\bm{\theta}^{-}$ & exponential moving average of  $\bm{\theta}$\\
    $\bm{\epsilon}_{\left\{ \cdot\right\}}$ & noise-prediction diffusion model\\
    $\bm{\vf}_{\left\{ \cdot\right\}}$ & consistency model\\
    $\vc_{\mathrm{txt}}$ & text prompt \\
    $\vc_{\mathrm{ctrl}}$ & new contional control \\
    $\vx$~/~$\vx_t$ & image~/~latent (noisy image)\\
\hline
\end{tabular}
\end{center}
\end{table}
\section{Introduction}
\label{sec:intro}
Consistency Models (CMs)~\cite{song2023consistency,song2023improved,luo2023latent,luo2023lcm} have emerged as a competitive family of generative models that can generate high-quality images in one or few steps. CMs can be distilled from a pre-tranined diffusion model or trained in isolation from data~\cite{song2023consistency,song2023improved}. Recently, latent consistency models (LCMs)~\cite{luo2023latent,luo2023lcm} have been successfully distilled from Stable Diffusion (SD)~\cite{rombach2022high}, achieving significant acceleration in the speed of text conditioned image generation. Compared with the glorious territory of diffusion models (DMs), an essential concern is whether there exists effective solutions for CMs to accommodate additional conditional controls. Inspired by the success of ControlNet~\cite{zhang2023adding} to text-to-image DMs, we consider to address this issue by training ControlNet for CMs.

In this technical report, we present three training strategies for ControlNet of CMs. Given the connection that CMs directly project any point of a probability flow ordinary differential equation (PF ODE) trajectory to data and DMs produce data by iterating an ODE solver along the PF ODE~\cite{song2020score}, we assume that the learned knowledge of ControlNet is (partially) transferable to CMs. Therefore, the first solution we try is to train ControlNet based on DMs and then directly apply the trained ControlNet to CMs. The advantage is that one can readily re-use the off-the-shelf ControlNet of DMs, but meanwhile at the cost of: i) sub-optimal performance. Due to the gap between CMs and DMs, the transfer may not be imperfect; ii) indirect training when adding new controls. That is, one has to utilize DMs as an agent to train a new ControlNet and then rely on the strong generalization ability of ControlNet to apply to CMs. 

Song et al.~\cite{song2023consistency,song2023improved} points out that CMs, as a family of generative models, can be trained in isolation from data by the consistency training technique. Inspired by this, we treat the pre-trained text-to-image CM and ControlNet as a new conditional CM with only ControlNet trainable.  Our second solution is to train the ControlNet using the consistency training. We find that ControlNet can also be successfully trained from scratch without reliance on DMs\footnote{Even if CMs may be established by consistency distillation from DMs.}. Building on the above two solutions, our third one involves training a multi-condition shared adapter to balance effectiveness and convenience.
Experiments on various conditions including edge, depth, human pose, low-resolution image and masked image suggest that:

\begin{itemize}
    \item ControlNet of DM can transfer high-level semantic controls to CM; however, it often fails to accomplish low-level fine controls;
    \item CM's ControlNet can be trained from scratch using the consistency training technique. Empirically, we can find that consistency training can accomplish more satisfactory conditional generation. 
    \item In addition, to mitigate the gap between DMs and CMs, we further propose to train a unified adapter with consistency training to facilitate to transfer DM's ControlNet; see examples in~\cref{fig:ctrl_maincompare}.
\end{itemize}

%% file: sec/2_method.tex
\section{Method}
Our method consists of four parts. First, we briefly describe how to train a text-to-image consistency model $\vf_{\bm{\theta}}\left(\vx_t, t; \vc_{\mathrm{txt}} \right)$ from a pre-trained text-to-image diffusion model $\bm{\epsilon}_{\bm{\phi}}\left(\vx_t, t; \vc_{\mathrm{txt}}\right)$ in \cref{sec:preparation}. We next introduces the first approach to train a ControlNet for a new condition $\vc_{\mathrm{ctrl}}$ by applying the diffusion model in \cref{sec:transfer_control}. Then, we propose to use the consistency training technique to train a ControlNet from scratch for the pre-trained text-to-image consistency model in \cref{sec:ct_control}. Last, we introduce a unified adapter that enables the rapid swift of multiple DMs-based ControlNets to CMs in \cref{sec:ct_adapter}. We summarize the meaning of symbols in \cref{tab:symbol} to help with readability.
\subsection{Preparation} \label{sec:preparation}
\begin{figure}[!htb]
    \centering
    \includegraphics[width=0.9\linewidth]{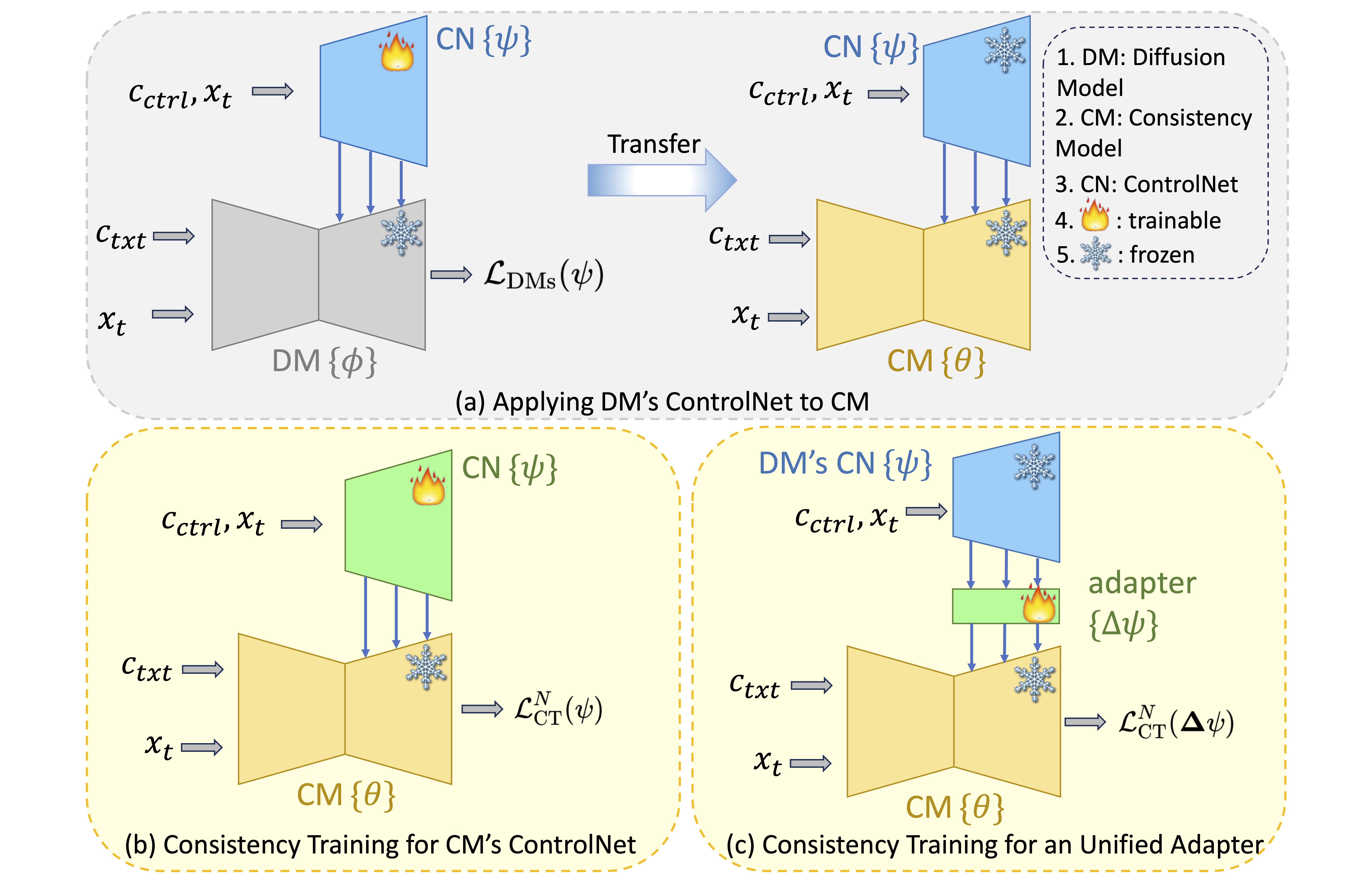}
    \caption{Overview of training strategies for ControlNet. (a) Training a ControlNet based on the text-to-image diffusion model (DM) and directly applying it to the text-to-image consistency model (CM); (b) consistency training for ControlNet based on the text-to-image consistency model; (c) consistency training for a unified adapter to utilize better transfer of DM's ControlNet.}
    \label{fig:method}
\end{figure}
The first step is to acquire a foundational text-to-image consistency model. Song et al.~\cite{song2023consistency} introduces two methods to train consistency models: consistency distillation from pre-trained text-to-image diffusion models or consistency training from data. Consistency distillation uses the pre-trained diffusion models to estimate score function (parameterized by $\bm{\phi}$). Given an arbitrary noisy latent $\left(\vx_{t_{n+1}}, t_{n+1}\right)$, an ODE solver is employed to estimate the adjacent latent with less noise, denoted as $\left(\hat{\vx}_{t_{n}}^{\bm{\phi}}, t_n\right)$. $\left\{\left(\vx_{t_{n+1}}, t_{n+1}\right), \left(\hat{\vx}_{t_{n}}^{\bm{\phi}}, t_n\right)\right\}$ belongs to the same PF ODE trajectory.  Then, consistency models can be trained by enforcing self-consistency property: the outputs are consistent for arbitrary pairs of $\left(\vx_t,t\right)$ of the same PF ODE trajectory. The final consistency distillation loss for the consistency model $\vf$ (parameterized by $\bm{\theta}$) is defined as
\begin{equation}
    \gL^N_{\mathrm{CD}}\left(\bm{\theta}, \bm{\theta}^-; \bm{\phi}\right) = \E_{\vx, \vx_{n+1}, \vc_{\mathrm{txt}}, n} \left[\lambda\left(t_n\right) d\left(\vf_{\bm{\theta}}\left( \vx_{t_{n+1}}, t_{n+1}; \vc_{\mathrm{txt}}\right), \vf_{\bm{\theta}^{-}}\left( \hat{\vx}_{t_{n}}^{\bm{\phi}}, t_{n}; \vc_{\mathrm{txt}}\right) \right)\right],
\end{equation}
where $\vx\sim \pdata$, $\vx_t\sim \gN\left( \sqrt{\alpha_t}\vx, (1-\alpha_t)\bm{I}\right)$ and  $n\sim \gU\left( \left[1, N-1 \right]\right)$. $\gU\left( \left[1, N-1 \right]\right)$ denotes the uniform distribution over $\left\{1,2,\ldots,N-1 \right\}$. According to the convention in Song et al.~\cite{song2023consistency}, $\vf_{\bm{\theta}^{-}}$ is the ``teacher network" and $\vf_{\bm{\theta}}$ is the ``student network" and $\bm{\theta}^{-} = \mathrm{stopgrad}\left(\mu \bm{\theta}^{-} + \left(1-\mu\right)\bm{\theta}\right)$. 

\subsection{Applying ControlNet of Text-to-Image Diffusion Models} \label{sec:transfer_control}
Given a pre-trained text-to-image diffusion model $\bm{\epsilon}_{\bm{\phi}}\left(\vx_t, t; \vc_{\mathrm{txt}}\right)$, to add a new control $\vc_{\mathrm{ctrl}}$, a ControlNet $\left\{ \bm{\psi}\right\}$ can be trained by minimizing $\gL\left(\bm{\psi}\right)$, where $\gL\left(\bm{\psi}\right)$ takes the form of 
\begin{align}
    \label{eq:train_diffuion}
    \gL_{\mathrm{DMs}}\left(\bm{\psi}\right)=\E_{\vx, \vc_{\mathrm{txt}}, \vc_{\mathrm{ctrl}}, \bm{\epsilon}}\left[\Vert
    \bm{\epsilon}- \bm{\epsilon}_{\left\{\bm{\phi}, \bm{\psi}\right\}}\left( \vx_{t},t; \vc_{\mathrm{txt}}, \vc_{\mathrm{ctrl}} \right) \Vert_2^2 \right].
\end{align}
In \cref{eq:train_diffuion}, $\vx\sim \pdata$ and $\bm{\epsilon}\sim \gN\left(\bm{0},\bm{I}\right)$. Suppose $\bm{\psi}^*=\argmin_{\bm{\psi}}\gL\left(\bm{\psi}\right) $, the trained ControlNet $\left\{\bm{\psi}^*\right\}$ is directly applied to the text-to-image consistency model. We assume that the learned knowledge to control image generation can be transferred to the text-to-image consistency model if the ControlNet generalizes well. Empirically, we find this approach can successfully transfer high-level semantic control but often generate unrealistic images. We suspect the sub-optimal performance is attributed to the gap between CMs and DMs.

\subsection{Consistency Training for ControlNet} \label{sec:ct_control}

Song et al.~\cite{song2023consistency,song2023improved} figures out that except consistency distillation from pretrained diffusion models, consistency models, as an independent class of generative models, can be trained from scratch using the consistency training technique. The core of the consistency training is to use an estimator of the score function:
\begin{align}
    \label{eq:estimator}
    \nabla \log p_t\left(\vx_t\right) &= \E\left[\nabla_{\vx_t} \log p(\vx_t|\vx) |\vx_t\right]\\
    &=-\E \left[ \frac{\vx_t-\sqrt{\alpha_t}\vx_t}{1-\alpha_t}\vert \vx_t\right],
\end{align}
where $\vx\sim \pdata$ and $\vx_t\sim \gN\left( \sqrt{\alpha_t}\vx, (1-\alpha_t)\bm{I}\right)$.  By a Monte Carlo estimation of~\cref{eq:estimator}, the resulting consistency training loss takes the mathematical form of
\begin{align}
    \label{eq:ctloss}
    \gL_{\mathrm{CT}}^N\left(\bm{\theta}\right)=\E_{\vx, \vx_t, n}\left[\lambda\left(t_n\right)d\left(\vf_{\bm{\theta}}\left( \vx_{t_{n+1}},t_{n+1} \right), \vf_{\bm{\theta}^{-}}\left( \vx_{t},t \right)\right) \right],
\end{align}
where the expectation is taken with respect to $\vx\sim \pdata$, $\vx_t\sim \gN\left( \sqrt{\alpha_t}\vx, (1-\alpha_t)\bm{I}\right)$ and  $n\sim \gU\left( \left[1, N-1 \right]\right)$. $\gU\left( \left[1, N-1 \right]\right)$ denotes the uniform distribution over $\left\{1,2,\ldots,N-1 \right\}$. 

To train a ControlNet for the pre-trained text-to-image consistency model (denoted as $\vf_{\bm{\theta}}\left( \vx_t, t; \vc_{\mathrm{txt}}\right)$ with the text prompt $\vc_{\mathrm{txt}}$), we consider to add a conditional control $\vc_{\mathrm{ctrl}}$ and define a new conditional consistency model $\vf_{\left\{\bm{\theta}, \bm{\psi}\right\}}\left( \vx_t, t; \vc_{\mathrm{txt}}, \vc_{\mathrm{ctrl}}\right)$ by integrating the trainable ControlNet $\left\{\bm{\psi}\right\}$ and the original frozen CM $\left\{\bm{\theta}\right\}$. The resulting training loss for ControlNet is
\begin{align}
    \label{eq:ctloss_cond}
    \gL_{\mathrm{CT}}^N\left(\bm{\psi}\right)=\E_{\vx, \vx_t, \vc_{\mathrm{txt}}, \vc_{\mathrm{ctrl}}, n}\left[\lambda\left(t_n\right)d\left(\vf_{\left\{\bm{\theta}, \bm{\psi}\right\}}\left( \vx_{t_{n+1}},t_{n+1}; \vc_{\mathrm{txt}}, \vc_{\mathrm{ctrl}}\right), \vf_{\left\{\bm{\theta}, \bm{\psi}\right\}^{-}}\left( \vx_{t_n},t_n; \vc_{\mathrm{txt}}, \vc_{\mathrm{ctrl}} \right)\right) \right].
\end{align}
Note that in \cref{eq:ctloss_cond}, only the ControlNet $\bm{\psi}$ is trainable. We simply set $\left\{\bm{\theta}, \bm{\psi}\right\}^{-}= \mathrm{stopgrad}\left(\left\{\bm{\theta}, \bm{\psi}\right\}\right)$ for the teacher model since recent research~\cite{song2023improved} reveals that omitting Exponential Moving Average (EMA) is both theoretically and practically beneficial for training consistency models. 

\subsection{Consistency Training for a Unified Adapter.} 
\label{sec:ct_adapter}
We find that DM's ControlNet can provide high-level conditional controls to CM. However, due to the presence of gap between CM and DM, the control is sub-optimal, i.e., it often causes unexpected deviation of image details and generate unrealistic images. To address this issue, we train a unified adapter to implement better adaption of DM's ControlNets $\left\{\bm{\psi_{1}, \ldots \psi_{K}}\right\}$ to CM using the consistency training technique. Formally, suppose the trainable parameter of the adapter is $\bm{\Delta\psi}$, the training loss for the adapter is:
\begin{align}
    \label{eq:ctloss_cond_adaptor}
    \gL_{\mathrm{CT}}^N\left(\bm{\Delta\psi}\right)=\E_{\vx, \vx_t, \vc_{\mathrm{txt}}, \vc_{\mathrm{ctrl}}, n,k}\left[\lambda\left(t_n\right)d\left(\vf_{\left\{\bm{\theta}, \bm{\psi_{k}}, \bm{\Delta\psi}\right\}}\left( \vx_{t_{n+1}},t_{n+1}; \vc_{\mathrm{txt}}, \vc_{\mathrm{ctrl}}\right), \vf_{\left\{\bm{\theta}, \bm{\psi_{k}}, \bm{\Delta\psi}\right\}^{-}}\left( \vx_{t_n},t_n; \vc_{\mathrm{txt}}, \vc_{\mathrm{ctrl}} \right)\right) \right],
\end{align}
where $k\sim \left[1, K \right]$ and K denotes the number of involved conditions. 
\cref{fig:ctrl_maincompare} shows that a lightweight adapter helps mitigate the gap and produces visually pleasing images.

%% file: sec/3_exp.tex
\section{Experiments}
\begin{figure}[!htb]
    \centering
    \includegraphics[width=\linewidth]{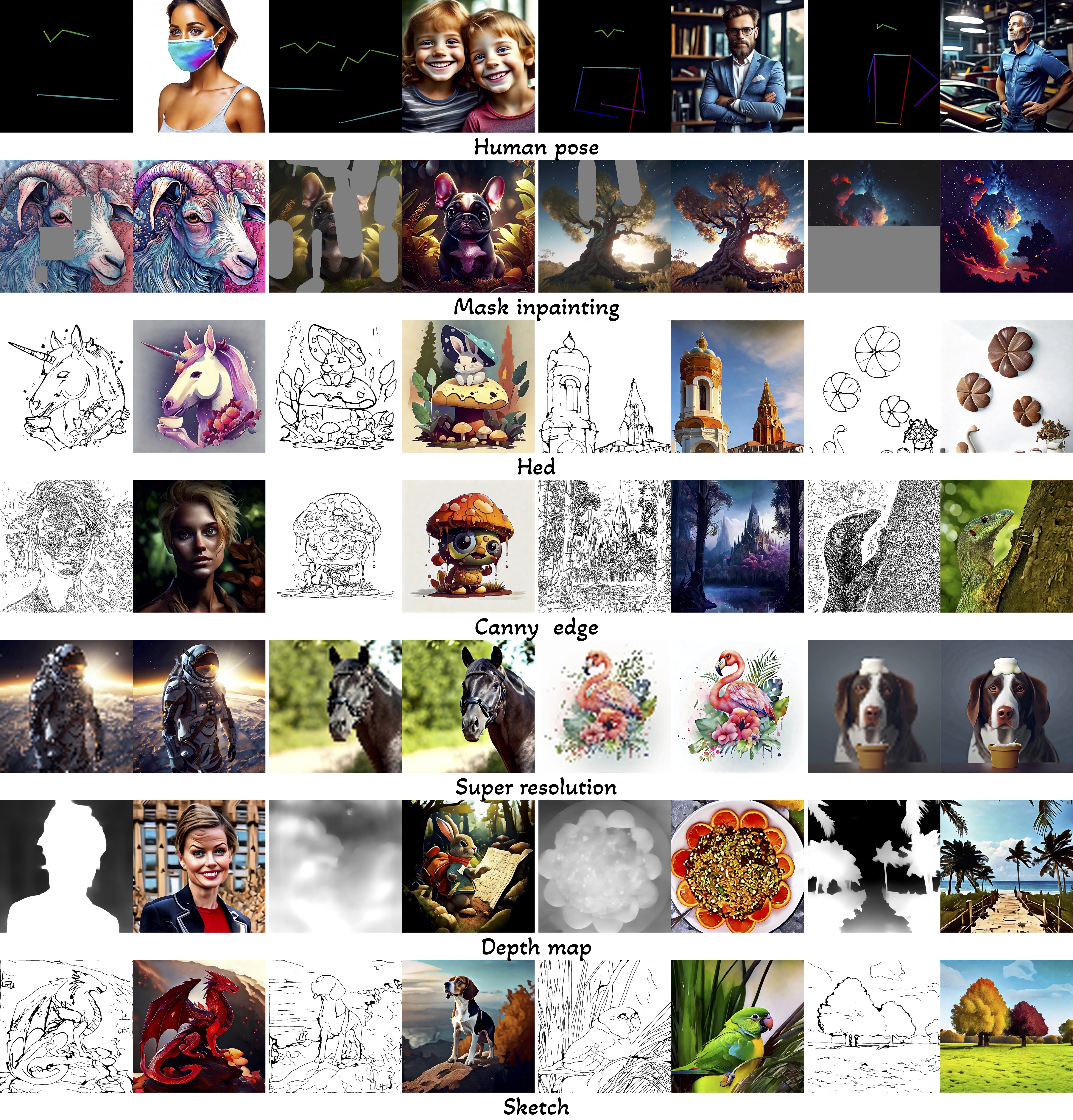}
    \caption{Images sampled by applying DM's ControlNet to CM at $1024$x$1024$ resolution. NFEs=$4$.}
    \label{fig:ctrl_transfer}
\end{figure}

\begin{figure}[!htb]
    \centering
    \includegraphics[width=\linewidth]{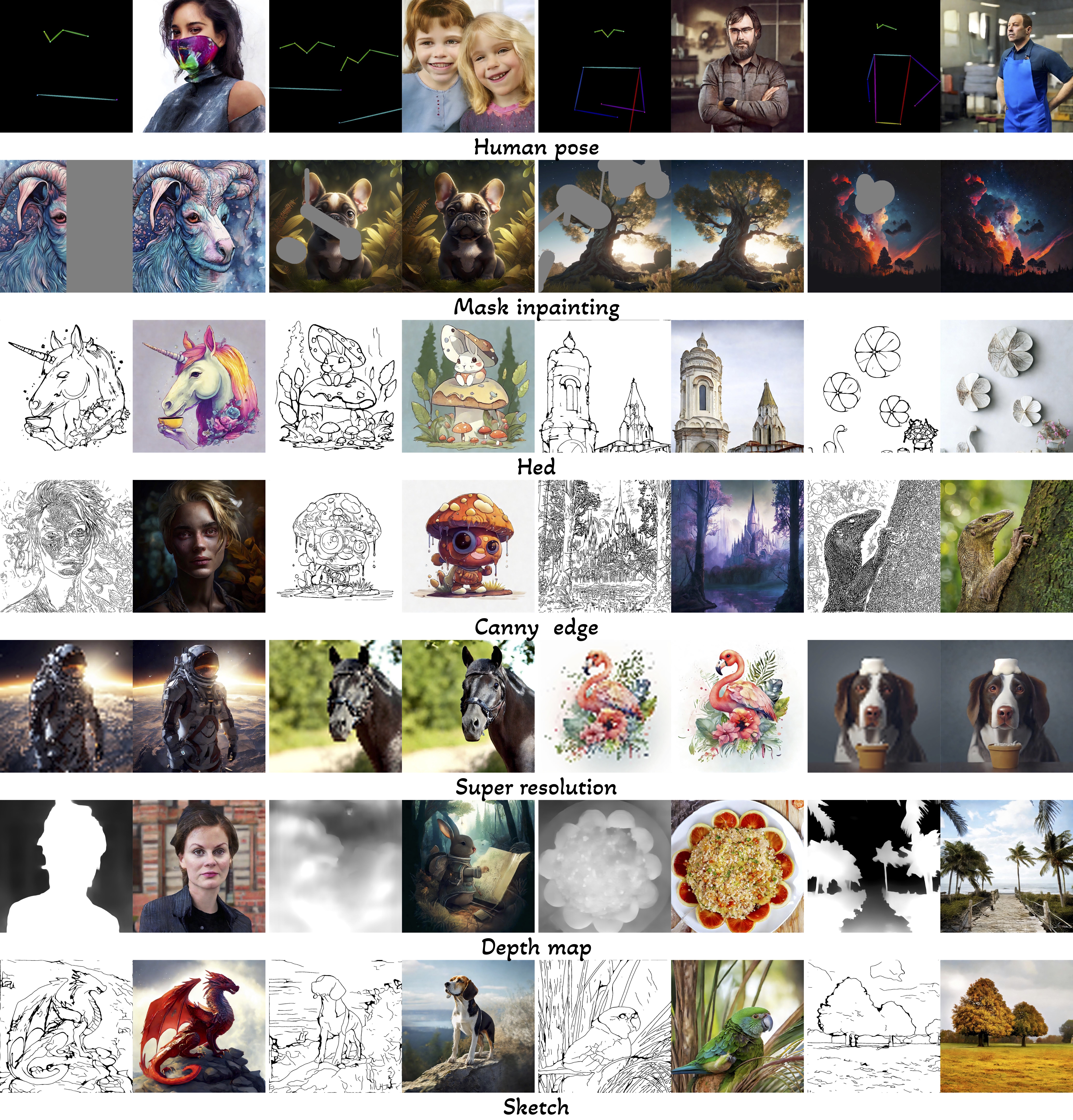}
    \caption{Visual results of consistency training at $1024$x$1024$ resolution. The conditions are the same with those in~\cref{fig:ctrl_transfer}. It can be observed that CM's ControlNet using consistency training can generate more visually pleasing images compared to DM's ControlNet. NFEs=$4$.}
    \label{fig:ctrl_reference}
\end{figure}

\begin{figure}[!htb]
    \centering
    \includegraphics[width=\linewidth]{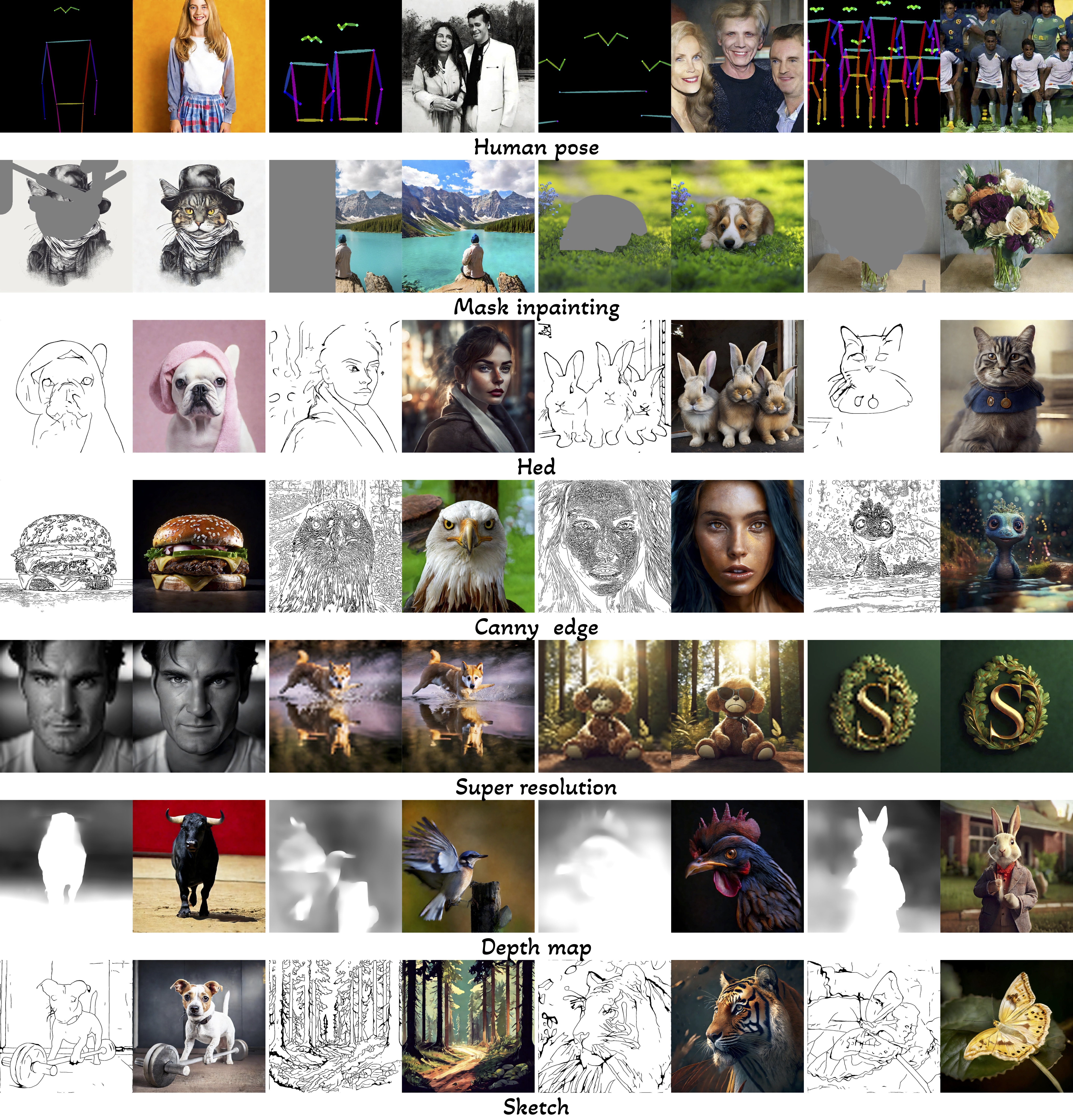}
    \caption{More visual results of CM's ControlNet using consistency training strategy at $1024$x$1024$ resolution. NFEs=$4$.}
    \label{fig:ctrl_ct}
\end{figure}

\begin{figure}[!htb]
    \centering
    \includegraphics[width=0.9\linewidth]{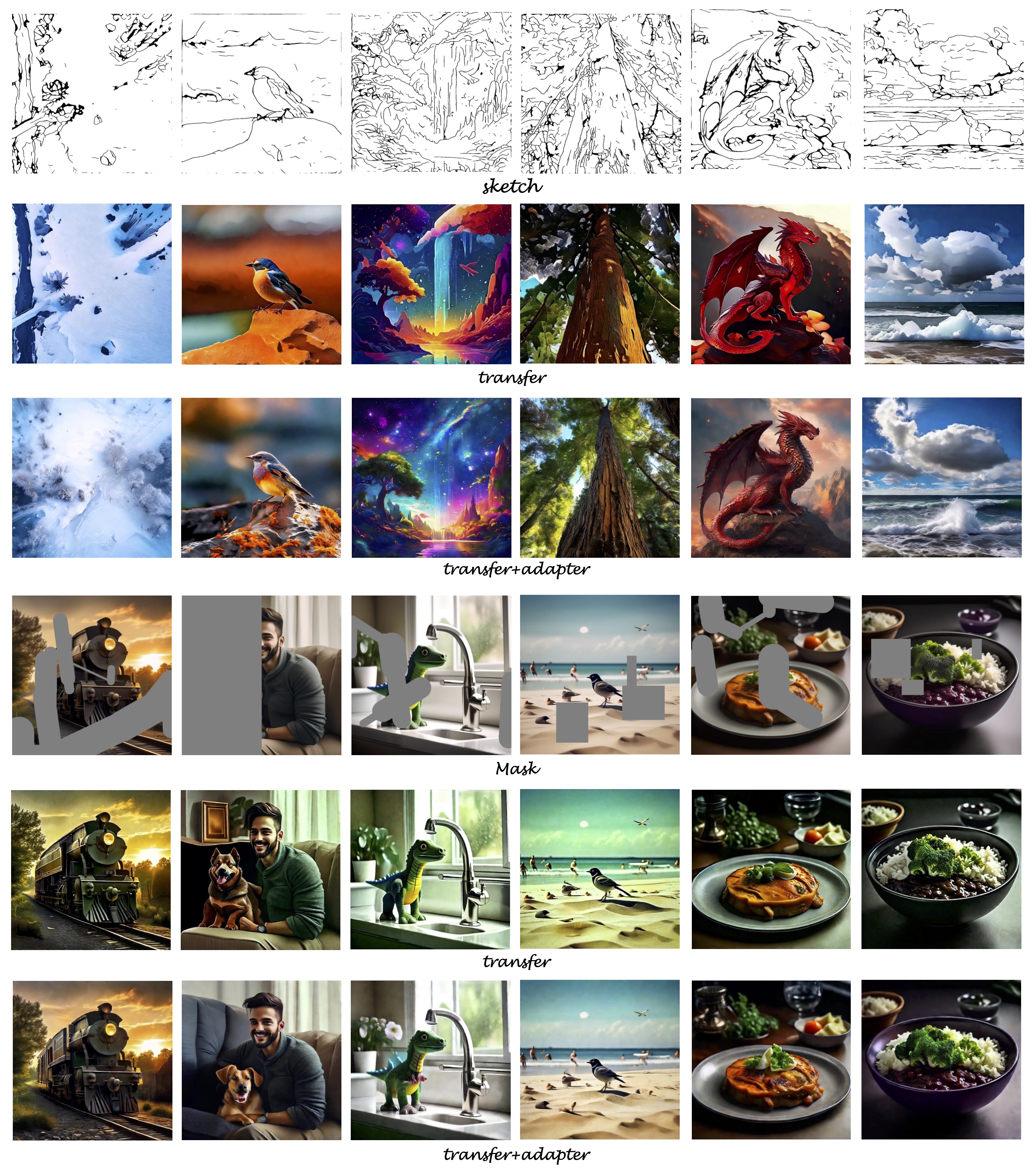}
    \caption{Visual results of  DM's ControlNet without/with a unified adapter at $1024$x$1024$ resolution. NFEs=$4$.}
    \label{fig:ctrl_adapter}
\end{figure}

\begin{figure}[!htb]
    \centering
    \includegraphics[width=0.9\linewidth]{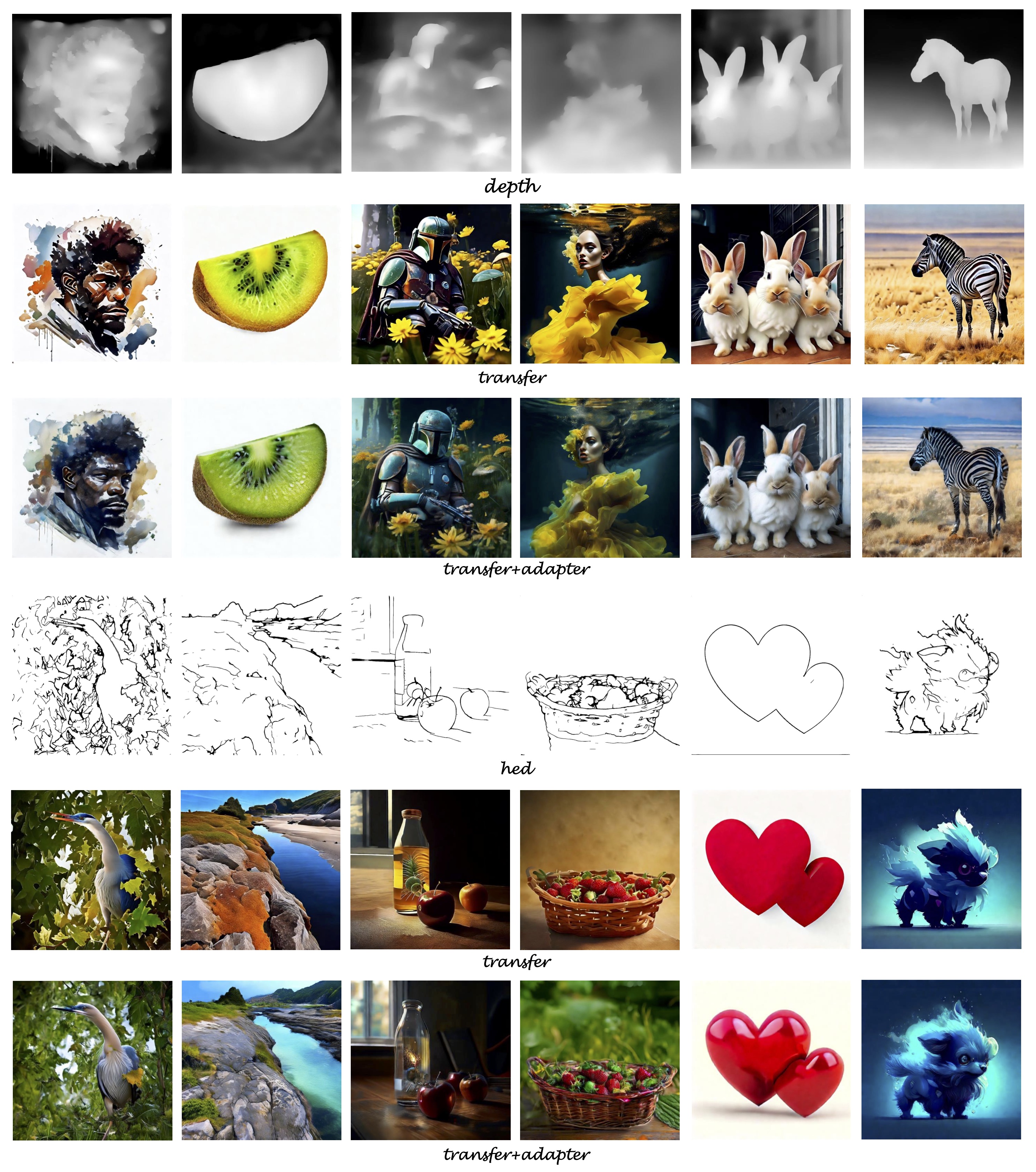}
    \caption{Visual results of  DM's ControlNet without/with a unified adapter on training-free conditions at $1024$x$1024$ resolution. NFEs=$4$.}
    \label{fig:ctrl_adapter2}
\end{figure}

\begin{figure}[!htb]
    \centering
    \includegraphics[width=0.9\linewidth]{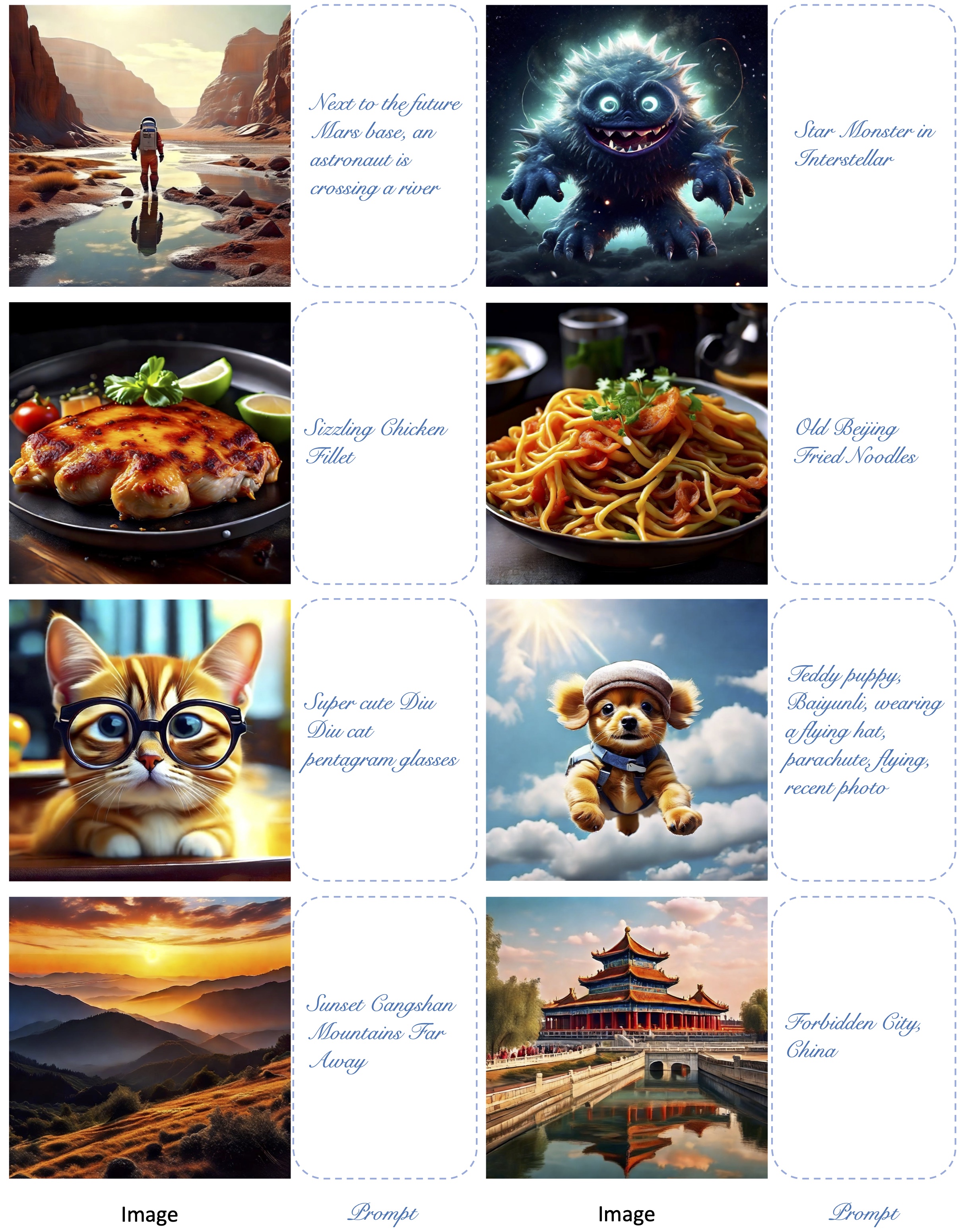}
    \caption{Images generated using our re-trained Text-to-Image CM with 4-step inference at $1024$x$1024$ resolution.}
    \label{fig:lcm_result}
\end{figure}
\subsection{Implementation Details}
\paragraph{Prepration.} To train the foundational consistency model, we set $\bm{\theta}^{-} = \mathrm{stopgrad}\left( \bm{\theta}\right)$, $N=200$, $\textrm{CFG}=5.0$, and $\lambda\left(t_n\right)=1.0$ for all $n\in\mathcal{U}([1, N-1])$. We enforce zero-terminal SNR~\cite{lin2023common} during training to align training with inference.
The distance function is chosen as the $\ell_1$ distance $d(\vx,\vy) = \Vert \vx-\vy \Vert_1$.
This training process costs about 160 A$100$ GPU days with 128 batch size.

\paragraph{Consistency Training.} To train ControlNets by consistency training, we set $\bm{\theta}^{-} = \mathrm{stopgrad}\left( \bm{\theta}\right)$, $N=100$, $\textrm{CFG}=5.0$, and $\lambda\left(t_n\right)=1.0$ for all $n\in\mathcal{U}([1, N-1])$.
The distance function is chosen as the $\ell_1$ distance $d(\vx,\vy) = \Vert \vx-\vy \Vert_1$.
We train on a combination of public datasets, including ImageNet$21$K~\cite{russakovsky2015imagenet}, WebVision~\cite{li2017webvision}, and a filter version of LAION dataset~\cite{schuhmann2022laion}. We elinimate duplicates, low resolution images, and images potentially contain harmful content from LAION dataset.
For each ControlNet, the training process costs about 160 A$100$ GPU days with $128$ batch size.
We utilize seven conditions in this work:
\begin{itemize}
    \item Sketch: we use a pre-trained edge detection model~\cite{su2021pixel} in combination with a simplification algorithm to extract sketches;
    \item Canny: a canny edge detector~\cite{canny1986computational} is employed to generate canny edges;
    \item Hed: a holistically-nested edge detection model~\cite{xie2015holistically} is utilized for the purpose;
    \item Depthmap: we employ the Midas~\cite{ranftl2020towards} for depth estimation;
    \item Mask: images are randomly masked. We use a $4$-channel representation, where the first 3 channels correspond to the masked RGB image, while the last channel corresponds to the binary mask;
    \item Pose: a pre-trained human-pose detection model~\cite{cao2017realtime} is employed to generate human skeleton labels;
    \item Super-resolution: we use a bicubic kernel to downscale the images by a factor of $16$ as the condition.
\end{itemize}
\subsection{Experimental Results}
\paragraph{Applying DM's ControlNet without Modification.} \cref{fig:ctrl_transfer} presents visual results of applying DM's ControlNet to CM. We can find that DM's ControlNet can deliver high-level controls to CM. Nevertheless, this approach often generates unrealistic images, \eg, Sketch in~\cref{fig:ctrl_transfer}. Moreover, DM's ControlNet of masked images causes obvious changes outsides the masked region (Mask inpainting in~\cref{fig:ctrl_transfer}). This sub-optimal control may explained that there exists the gap between CM and DM, which further causes imperfect adaptation of DM's ControlNet to CM. 

\paragraph{Consistency Training for CM's ControlNet.} For fair comparison, \cref{fig:ctrl_reference} shows corresponding visual results of consistency training for ControlNet.  We can find that consistency training directly based on CM can generate more realistic images. Therefore, we can conclude that consistency training offers a way to train the customized ControlNet for CMs. More generative results can be found in \cref{fig:ctrl_ct}.
\paragraph{Transferring DM's ControlNet with a Unified Adapter.} When compared to direct transfer method, a unified adapter trained under five conditions (\textit{i.e.,} sketch, canny, mask, pose and super-resolution) enhances the visual quality of both in-context images (\textit{e.g.,} sketch and mask conditions in \cref{fig:ctrl_adapter}) and training-free conditions (\textit{i.e.,} depthmap and hed conditions in \cref{fig:ctrl_adapter2}), showing promising prospects.

\paragraph{Real-time CM Generation.} To comprehensively evaluate the quality of images generated under the aforementioned conditions, \cref{fig:lcm_result} presents the effects of our re-trained text-to-image CM model during four-step inference.

%% file: sec/4_relate_work.tex
\section{Related Work}
\paragraph{Real-time Generation} We briefly review recent advancements in accelerating DMs for real-time generation. 
Progressive distillation~\cite{salimans2022progressive} and guidance distillation~\cite{meng2023distillation} introduce a method to distill knowledge from a trained deterministic diffusion sampler, which involves multiple sampling steps, into a more efficient diffusion model that requires only half the number of sampling steps.
InstaFlow~\cite{liu2023instaflow} turns SD into an ultra-fast one-step model by optimizing transport cost and distillation.
Consistency Models (CMs)~\cite{song2023consistency, song2023improved} propose a new class of generative models by enforcing self-consistency along a PF ODE trajectory. Latent Consistency Models (LCMs)~\cite{luo2023latent} and LCM LoRA~\cite{luo2023lcm} extend CMs to enable large-scale text-to-image generation.
There are also several approaches that utilize adversarial training to enhance the distillation process, such as UFOGen~\cite{xu2023ufogen}, CTM~\cite{kim2023consistency}, and ADD~\cite{sauer2023adversarial}.

\paragraph{Controllable Generation} 
ControlNet~\cite{zhang2023adding} leverages both visual and text conditions, resulting in impressive controllable image generation.
Composer~\cite{huang2023composer} explores the integration of multiple distinct control signals along with textual descriptions, training the model from scratch on datasets of billions of samples.
UniControl~\cite{qin2023unicontrol} and Uni-ControlNet~\cite{zhao2023uni} not only enable composable control but also handle various conditions within a single model. They are also capable of achieving zero-shot learning on previously unseen tasks.
There are also several customized methods, such as DreamBooth~\cite{dreambooth}, Custom Diffusion~\cite{cosdiff}, Cones~\cite{cones,cones2},  and Anydoor~\cite{chen2023anydoor}, that cater to user-specific controls and requirements.

%% file: sec/5_conclus.tex
\section{Conclusion}
We study three solutions of adding conditional controls to text-to-image consistency models. The first solution directly involves a pre-trained ControlNet based on text-to-image diffusion model into text-to-image consistency models, showing sub-optimal performance. The second solution is to treat the text-to-image consistency model as an independent generative model and train a customized ControlNet using the consistency training technique, exhibiting exceptional control and performance. Furthermore, considering the strong correlation between DMs and CMs, we introduce a unified adapter into the third solution to mitigate the condition-shared gap, resulting in promising performance.

\clearpage